\documentclass[runningheads]{llncs}
\usepackage{graphicx}

\usepackage{siunitx}
\usepackage{caption}
\usepackage{subcaption}
\usepackage{breqn}

\usepackage{svg}
\usepackage{comment}
\usepackage{amsmath,amssymb} 
\usepackage{color}

\usepackage{tikz}
\usepackage{pgfplots}
\pgfplotsset{compat=newest}
\usepgfplotslibrary{units}

\usepackage{caption}
\usepackage{hyperref}

\usepackage[utf8]{inputenc}
\usepackage{algorithm}
\usepackage[noend]{algpseudocode}

\usepackage[accsupp]{axessibility}  
\newcommand{\argmin}{\mathop{\mathrm{argmin}}\nolimits} 
\DeclareUnicodeCharacter{2212}{\textminus}

\begin{document}
\pagestyle{headings}
\mainmatter
\def\ECCVSubNumber{3175}  

\title{Plausibility Verification For 3D Object Detectors Using Energy-Based Optimization} 


\titlerunning{Plausibility checks For 3D Obj. Detectors}
%
\author{Abhishek Vivekanandan\inst{1} \and
Niels Maier\inst{2} \and
J. Marius Zöllner\inst{1,2}}
\authorrunning{Vivekanandan et al.}
%
\institute{ FZI Research Center for Information Technology, Germany \and
Karlsruhe Institute of Technology, Germany \\
\email{vivekana@fzi.de}}
\maketitle

\begin{abstract}
Environmental perception obtained via object detectors have no predictable safety layer encoded into their model schema, which creates the question of trustworthiness about the system's prediction. As can be seen from recent adversarial attacks, most of the current object detection networks are vulnerable to input tampering, which in the real world could compromise the safety of autonomous vehicles. The problem would be amplified even more when uncertainty errors could not propagate into the submodules, if these are not a part of the end-to-end system design. To address these concerns, a parallel module which verifies the predictions of the object proposals coming out of Deep Neural Networks are required. This work aims to verify 3D object proposals from MonoRUn model by proposing a plausibility framework that leverages cross sensor streams to reduce false positives. The verification metric being proposed uses prior knowledge in the form of four different energy functions, each utilizing a certain prior to output an energy value leading to a plausibility justification for the hypothesis under consideration. We also employ a novel two-step schema to improve the optimization of the composite energy function representing the energy model.
\keywords{Plausibility, Safety, 3D Object detection, SOTIF}
\end{abstract}

\section{Introduction}
Adversarial attacks on object detection networks are making the real-world deployment of Neural Networks (NN) susceptible to safety violations, hindering the approval and conformance of vehicles to SOTIF standards. This is attributed mainly to the black box nature of NN themselves. Often the perception module, with the object detector at its core, plays a key part in situations where these errors occur. An object detector's (OD) failure to perceive an object (e.g., another car or a pedestrian crossing the street) can immediately result in an unsafe situation both for the vehicle's passengers and other traffic participants.\\
However, detections which are misclassified or falsely proposed by the OD also constitutes considerable risk under the Operational Design Domain (ODD) of an automated vehicle. These artifacts are called as \textit{ghost detections} or \textit{false positives} often appear due to perception gaps or sensor noises. False positives not only cause sudden jerks, contributing to an uncomfortable driving experience, but could also lead to rear end collisions when braking is applied without the need for it.
Motivated by the dangers and risk posed, we propose to develop a parallel checker module following the architecture design of Run-Time Assurance (RTA)~\cite{Cofer2020} tests. Through experiments, we extensively show that our module checks for the plausibility of an object (in this work we chose the category car) using combinations of energy functions, thereby significantly reducing the number of False Positives relative to the base NN. 
The energy functions are made up of simple priors, which conforms to our definition of world knowledge \cite{VonRueden2019}\cite{vivekana}.
The base network under consideration is MonoRUn~\cite{monoRUNhuangyuyao} which uses RGB images to predict 3D Bounding Boxes defining the position and orientation of an object, forming an initial hypothesis which needs to be checked for.
As shown in Fig. \ref{fig:General_architecture}, the checker modules (marked through dashed rectangle boxes) uses raw LiDAR point clouds and Camera streams with the assumption that the sensors are synchronized accurately to provide valuable environmental information to different parts of the module. \\
In summary, we: 1) Design a parallel checker module for plausibility checks for the outputs of a 3D-OD network. 2) Propose a novel two-step optimization schema for composite energy function, which depends on 3D shape priors. 3) Developed computationally light rendering module to obtain 2D segmentation masks from the optimized 3D shape priors represented through the notation $(y^*,z^*)$. 4) Finally, we propose a simple empirically evaluated threshold based False Positive filter with the help of an energy-based model.

\section{Related work}
Several works aim to verify the detected objects' existence in a fusion system.
Often, the Dempster–Shafer theory of evidence (DST) \cite{Shafer1976} is used to implement and combine plausibility features on various system levels.
In \cite{Aeberhard2011}, authors generate object existence probability using a high-level architecture defining different sensor fusion for their autonomous test vehicle.
Each individual sensor assigns existence probabilities to the objects it detects. 
By following the DST rules of combination, existence probabilities of each sensor are merged by constructing basic belief assignments.
In their work \cite{Geissler2020}, the authors present a similar fusion system applied to the detection of cars from a roadside sensor infrastructure at a highway section consisting of various radar sensors. 
Different plausibility checks for individual sensors are encoded as basic belief assignments through \textit{a priori} assumptions on geometric constraints (such as a cameras' field of view or occlusions) and parameters such as the trust into a sensors' performance, which yields a Bayesian like probability of an object's existence.
The authors from \cite{Khesbak2021} present a different approach for plausibility evaluation about an object's existence by employing a serial implementation checks for the consensus between two detectors from different sensor streams. A simple measurement of Latency-based threshold against a pre-defined distance threshold checks for implausibility. Works from \cite{maagKira,rottmannMatthias}, propose False Negatives and False positive reduction algorithm on semantic segmentation tasks. 
Energy-based models (EBM) have played an important part throughout the history of pre-modern machine learning. In \cite{lecun06}, LeCun et al. gives an extensive tutorial and introduction to energy-based learning methods.
They describe the concept of EBMs as capturing dependencies between variables by associating a scalar valued energy to each configuration of the variable. Energy functions are a way to encode certain priors over a set of variable states (defined through properties and entities of a system), which yields a net-zero energy value for a perfect compatibility between variables during  inference and high-energy value for incompatible variables.
In \cite{Osadchy2006}, for the discriminative task of object detection, the pose is modelled as a latent variable, and all parameters are obtained by training with a contrastive loss function from a facial pose dataset by minimizing an energy function.\\
Engelmann et al. \cite{Engelmann2016} used a set of hand designed energy functions together with their proposed shape manifold for object segmentation task as well as pose and shape estimation.
For the energy functions, they combine a Chamfer-Distance Energy, measuring the distance from the stereo points to the object's surface with some prior constraints, punishing deviations of the shape from the mean pose and the object's height over ground.
From the works of \cite{Prisacariu2012.3305}, \cite{Rao2017}, \cite{Wang2020}, they apply latent shape space approaches to recover 3D shapes of a car using EBMs.

Differing from the previous presented data-driven approaches, Gustafsson et al. \cite{Gustafsson2020} aim to refine 3D object detection without relying on shape priors.
Their work builds upon their previous work \cite{Gustafsson2019}, in which confidence-based regression is performed on 2D detection results using an energy-based model. 
In their more recent work, they extend this idea by proposing to use a Deep Neural Network (DNN) to train a conditional energy-based model for probabilistic regression of 3D bounding boxes on point cloud data.

In comparison to the presented approaches here, we propose a novel approach of encoding priors into an energy-based system, from which we measure compatibility by forming a decision rule through energy value thresholds ensuring the plausibility of an object.
\section{Concepts and Proposed method}
\begin{figure*}[htb]
	\centering
    \includegraphics[width=\textwidth]{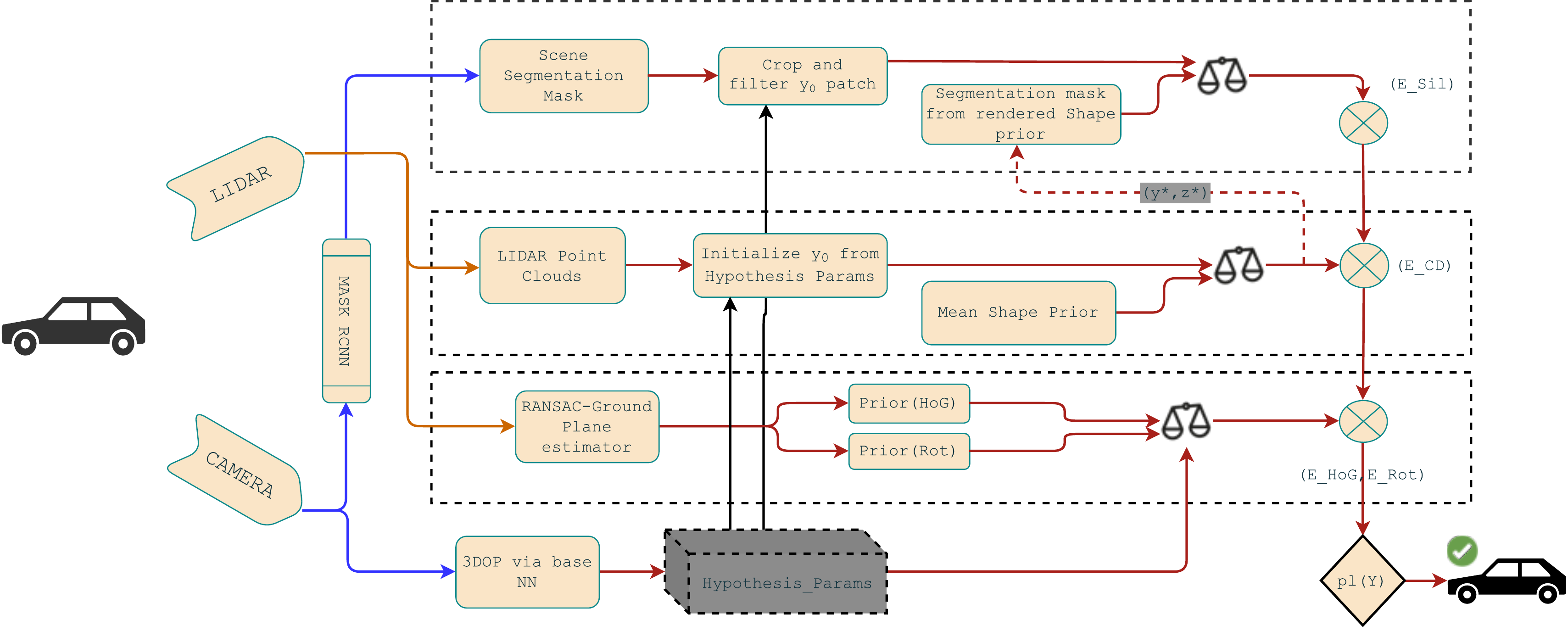}
    \caption{General architecture describing the flow of energy values which are combined to argue about an object's presence. $y_{0}$ represents the initial hypothesis. LiDAR and Camera inputs are primarily used to provide cross sensor data streams. Top: $E_{Sil}$, compares segmentation mask from MaskRCNN \cite{He2017} and rendered segmentation mask from the optimized outputs of CD energy function. Middle: $E_{CD}$, compares and optimizes a mean 3D shape prior with raw segmented point clouds within the hypothesis space. Bottom: $E_{HoG},E_{Rot}$, compares ground estimates from a RANSAC regressor with the height and rotation of an initial hypothesis}
    \label{fig:General_architecture}
\end{figure*}
The main idea is to encode prior knowledge into the energy function such that a comparison could be made between the priors ($\pmb{X} \in \mathcal{X}_i$) and the predictions from the 3D object detector ($\pmb{y} \in \mathcal{Y}$). 
The predictions from a 3D-OD from here on would be expressed as a hypothesis, as we aim to justify whether there exists an object inside the proposed space. Individual energy functions encode a specific prior knowledge and map the observed compatibility to a single scalar value. In mathematical terms, this can be expressed in a general definition of the energy functions as, $E: \mathcal{X}_i \times \mathcal{Y} \to \mathbb{R}$ and $E(\pmb{X}, \pmb{y}) \geq 0 ~ \forall \pmb{X},\pmb{y} \in \mathcal{X}_i \times \mathcal{Y}$.\\ 
The energy functions output energy values which exhibit the property of compatibility. For a perfect compatibility, $E =E^*=0$; while for any deviations, $E>0$. In this paper, we use four different energy functions, where each of the function constitutes a certain prior. 
The energy function $E_{Sil}$ measures silhouette alignment; $E_{CD}$, measures the alignment of a point cloud and the remaining two, $E_{HoG}$ and $E_{Rot}$ are based on ground estimates which measures the height over ground and rotational consistency of the bounding boxes.
The first two energy functions depend on additional prior knowledge in the form of 3D shape priors represented as a \textit{Truncated Signed Distance Function} (TSDF) through CAD models based on ShapeNet \cite{Chang2015}.
The two energy functions based on the ground estimate do not depend on any additional input, and are referred to as energy priors throughout this work.
Each energy function can be interpreted as an expert, having expertise about one aspect that contributes to the plausibility measure.
To be able to obtain feasible plausibility evaluations, the individual energy functions can be combined into an energy-based model defined as:
\begin{equation}
    E = \alpha_0 E_{Sil} + \alpha_1 E_{CD} + \alpha_2 E_{HoG} + \alpha_3 E_{Rot}
\end{equation}
The concatenation of such uncalibrated experts are called as product-of-experts \cite{Hinton99}.

\subsection{Chamfer Distance Energy Function}
Inspired from the works of \cite{Engelmann2016} and \cite{Rao2017}, we construct our first energy function based on the \textit{Chamfer Distance} (CD) to evaluate the alignment/compatibility of the shape priors and acquired point cloud data through optimization. 
The \textit{Chamfer Distance} is the summation of the closest point distances between two sets of points. For two point clouds $\mathcal{A}$ and $\mathcal{B} \in \mathbb{R}^3 $:
\begin{equation}
    CD(\mathcal{A}, \mathcal{B}) = \sum_{a \in \mathcal{A}} \min_{b \in \mathcal{B}} \Vert a - b \Vert_2^2 + \sum_{b \in \mathcal{B}} \min_{a \in \mathcal{A}} \Vert a - b \Vert_2^2
\end{equation}
This distance measure as such satisfies the requirements of an energy function.
However, \cite{Engelmann2016} proposed further modification to the existing energy function to better handle the robustness and to make it compatible with second order optimizations.
To reduce the effect of outliers points during optimization and also for a better signal-to-noise ratio; the \textit{ Huber loss function} $\rho: \mathbb{R} \to \mathbb{R}$ is applied to the squared TSDF values. In order to make the function independent of the point cloud size, we take the mean of all points (summed distance of the point's distance values are divided by the total number of points in the point cloud $N= \mid \mathcal{X}_{PC}\mid$).
With these considerations, the Chamfer distance energy function measures the compatibility between a shape prior $\pmb{\Phi} = \pmb{\Phi}(\pmb{\tau}, \pmb{z})$ (parameterized by the 3D pose $\pmb{\tau}$ and shape weights $\pmb{z}$) and the raw point clouds inside the hypothesis space. Such a measure, throughout this work, is defined as:
\begin{equation}
    E_{CD}(X_{PC},\pmb{\Phi}) = \frac{1}{N} \sum_{\pmb{x}_i \in \mathcal{X}_{pc}} \rho(\Phi(\pmb{x}_i, \pmb{z})^2)\\
    \ni \\ \rho(x) = \begin{cases} x &, \, x \leq \varepsilon \\ 2 \sqrt{x} - \varepsilon  &, \, \text{otherwise} \end{cases} 
\end{equation}\\
As our main goal is to verify that an object of class car is present inside the proposals from the 3D-OD, we fetch this information via a query (Fig. \ref{fig:General_architecture} black solid arrows from \textit{Hypothesis\_Params} block). 
A query contains pose parameters representing the hypothesis's bounding box coordinates 
along with the mean shape manifold ($\Phi_{mean}$, obtained from the set of objects from ShapeNet) forming the initial hypothesis ($\pmb{y}_0$).


\subsection{Silhouette Alignment Energy function (SAEF)}
The SAEF measures the consistency between the silhouette alignment of a given segmentation mask and a rendered silhouette mask of an object hypothesis from the projection function.
While the matched segmentation mask of an object $M$ is obtained from the instance segmentation network (MaskRCNN), the object hypothesis silhouette is obtained by projecting the optimized 3D TSDF shape prior $\pmb{\Phi}$ into the image space based on its current shape and pose estimate. As shown in the Fig. \ref{fig:General_architecture}, dotted red line from the optimized CD energy function going into the \textit{segmentation mask creation} block.
This rendering of the shape prior is expressed through the projection function $\pi(\pmb{\Phi}, \pmb{p}): \pmb{\Phi}, \pmb{p} \mapsto (0, 1)$, assigning each pixel $\pmb{p}$ a value close to $1$ inside the object and close to zero outside the object.  

\begin{equation}
    E_{Sil}(\pmb{\Phi}) = \frac{1}{\mid{\Omega}\mid} \sum_{\pmb{p} \in \Omega} r_{sil}(\pmb{p}, \pmb{\Phi})
\end{equation}
where $\Omega$ is the set of pixels (i.e., the region of interest in the image) and $r_{Sil}$ is the residual comparing the segmentation masks per pixel:
\begin{equation}
    r_{Sil}(\pmb{p}, \pmb{\Phi}) = - \ln (p_{fg}(\pmb{p})\pi(\pmb{\Phi}, \pmb{p}) + p_{bg}(\pmb{p})(1-\pi(\pmb{\Phi},\pmb{p}))
\end{equation}

Here $p_{fg}$ and $p_{bg}$ are the foreground and background probabilities of each pixel derived from the segmentation mask $M$.
The residual function emits large positive values if there exist inconsistencies between the segmentation mask and object hypothesis silhouette mask. On the other hand, for consistencies' between the pixels, the $\ln()$ function becomes close to $1$, resulting in a residual value close to $0$.

\textbf{Differentiable Rendering,}
inspired from the works of \cite{Prisacariu2012.3305}~\cite{Rao2017},  about silhouette masking, we design our projection function $\pi(\pmb{\Phi}, \pmb{p})$ through the following equation.
\begin{equation}
    \pi(\pmb{\Phi}, \pmb{p}) = 1 - \argmin_ {\Phi(\pmb{x}_i^{o}) \forall \pmb{x}_i^{o} \in \mathcal{X}_{ray}^{p}} \frac{1}{\exp^{\Phi(\pmb{x}_i^{o}) \xi} + 1}
\end{equation}

where $\mathcal{X}_{ray}^p$ are 3D points sampled along a ray that is cast from the camera center through the pixel $\pmb{p}$ in the pinhole camera model. 
The super-script $^o$ denotes the transformation from camera coordinate system into object coordinate system for evaluation in the TSDF shape grid. The function inside the product is the sigmoid function, where $\xi$ controls the sharpness of the inflection, translating to the smoothness of the projection contours.

The idea of this projection function is, that if a point sample along the ray cast through a pixel falls into the object shape in the 3D space, the point will be assigned a negative value through the shape's TSDF. 
For negative values, the sigmoid function, acting as a continuous and steadily differentiable approximation of the \textit{Heaviside step function}, will take a value close to $0$.
Points falling outside the shape will be assigned positive signed distance values, leading to values close to $1$ in the sigmoid function.

Through this definition of the projection function, the rendering process of the shape prior silhouette masks becomes a fully differentiable function.
This allows to analytically calculate desired gradients and Jacobians that can be used for optimizing the energy function.
\begin{figure}[ht]
	\centering
    \begin{subfigure}[b]{0.49\columnwidth}
        {\includegraphics[width=1.0\columnwidth]{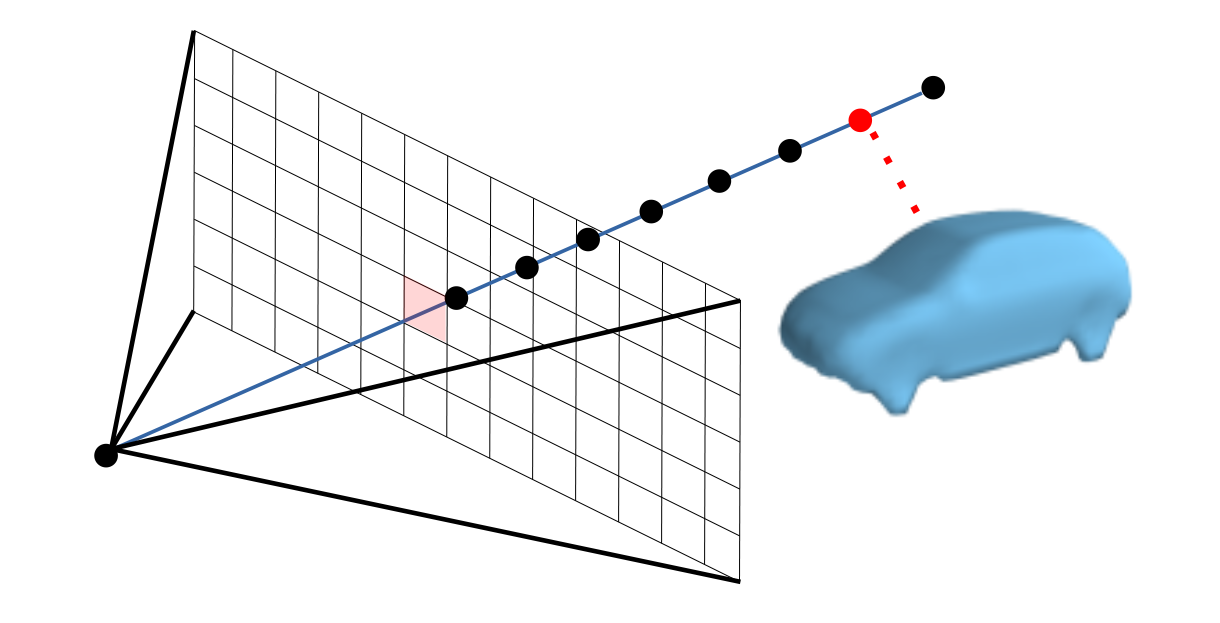}}
        \caption{Differentiable rendering for a single pixel}
        \label{fig:diff-render-scheme}
    \end{subfigure}
    \begin{subfigure}[b]{0.49\columnwidth}
        {\includegraphics[width=1.0\columnwidth]{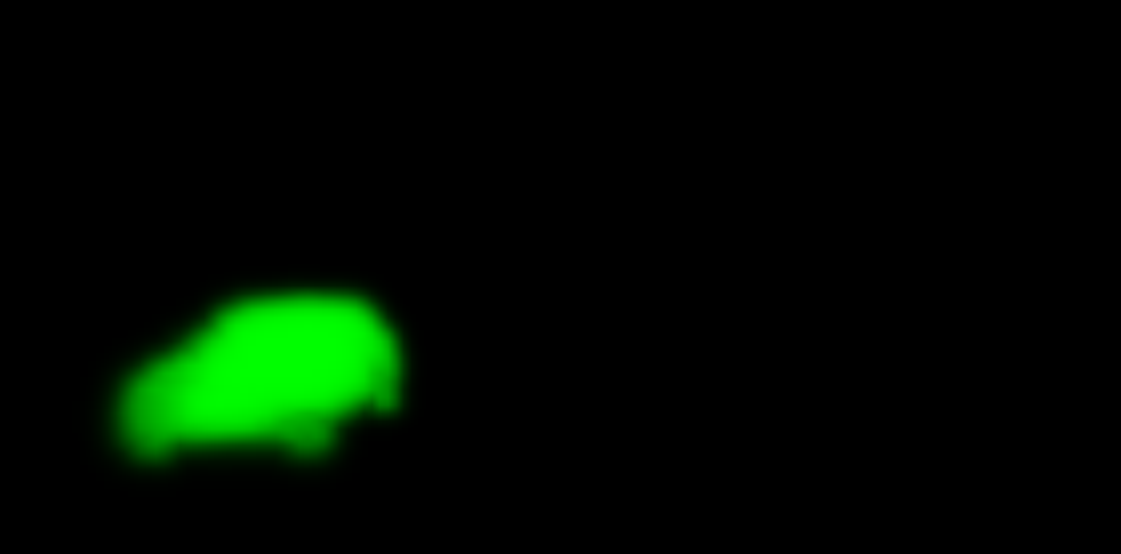}}
        \caption{Rendered Mask from optimized shape prior}
        \label{fig:diff-render-example}
    \end{subfigure}
    \caption{Schematic example for the differentiable rendering process obtained from the modified projection function.}
    \label{fig:diff-render}
\end{figure}
By taking into consideration only the point along each ray with the minimal signed distance value, the number of points in the further evaluation is reduced significantly. This helps us to significantly speed up the rendering process.
A schematic of this process for one pixel is shown in Fig. \ref{fig:diff-render-scheme}.
The red-point is the sampled point along the cast ray that is closest to the shape and therefore determines the pixel value.
The example rendered silhouette mask using this approach shown in Fig. \ref{fig:diff-render-example} establishes how through this modification, the purpose of the projection function is preserved when compared with \cite{Rao2017}.
The point with the minimal TSDF value sampled along the ray for a pixel containing the objects' projection will still be a negative value, such that the projection function returns a value close to $1$ (green in Fig. \ref{fig:diff-render-example}).
Similar, for those pixels that do not contain the object's projection, the function will still return a value close to $0$.

\subsection{Height over Ground Energy Function}
\label{sec:hog-prior}
One simple requirement for a plausible hypothesis is, for the object to be on the ground.
Especially for detections of the class \textit{Car,} this is a strong requirement.
Through ground plane estimation (RANSAC regressor), the ground level of each coordinate in the x-y plane of a scene can be approximated through the function:
\begin{equation}
    g(\pmb{t}(x, y)) = \frac{1}{c} (d - ax - by)
\end{equation}
where $\pmb{t} \in \mathbb{R}^3$ is the pose of an object hypothesis in the ego-vehicle coordinate system.
This allows to calculate the height over ground of an object hypothesis as
\begin{equation}
    d_{HoG}(\pmb{t}, h) = \pmb{t}(z) - \frac{h}{2} - g(\pmb{t}(x, y))
\end{equation}
where $h$ denotes the hypothesis height, accounting for the offset in height of the 3D pose parametrization chosen as the 3D bounding box center of the object.\\
Using the height over ground distance, an energy function encoding this requirement can be formulated in a parabolic form as proposed in \cite{Wang2020} and \cite{Engelmann2016}:
\begin{equation}
    E_{HoG}(\pmb{y}) = (d_{HoG}(\pmb{t}, h))^2 = (\pmb{t}(z) - \frac{h}{2} - g(\pmb{t}(x,y)))^2
\end{equation}

For objects being close to the ground plane, the energy function will be close to $0$, while deviations are punished quadratically.

\subsection{Rotation Consistent Energy Function}
Similarly to the height-over-ground energy prior, another prior assumption for the orientation of an object detection hypothesis based on a ground estimate can be formulated as an energy prior.
Following a similar line of reasoning for objects of class \textit{Car} to be able to touch the ground, this requirement specifies that they should touch the ground with the wheels.
This requirement of the hypothesis's orientation to align with the ground can be formulated as the hypothesis's z-axis being parallel to the ground normal vector $\pmb{n}_g$. 
Given the orientation of the hypothesis w.r.t. the ego vehicle coordinate system through the rotation matrix $R$ and the estimated ground plane normal, $\vec{n}_0 = [a, b, c]^T$ gives the scalar product that can be used to evaluate the alignment of the hypothesis axis and the normal vector.\\
The scalar product of two vectors has the property of yielding values close to $0$ for orthogonal vectors, while for two normal parallel vectors it  evaluates to $1$. The rotation energy prior can be defined as:
\begin{equation}
    E_{rot} = (1 - (R\cdot (0, 0, 1)^T)^T \cdot n_g)^2
\end{equation}

The dot product of the rotation matrix $R$ and the vector $(0, 0, 1)^T$ is equal to taking the last row of the rotation matrix as it is describing the object's z-axis rotated to the ego-vehicle coordinate system.

\section{Method}
The energy-based model now describes an energy surface, on which each point in the object hypothesis space $\mathcal{Y}$ (parameterized by object 3D pose and shape prior primary components) is assigned with a value $E \geq 0$ that directly expresses the compatibility of the hypothesis through the observed raw data under the assumption of a certain shape prior.  
Optimal plausible hypotheses can be understood as being close to a local minimum (considering noise, etc.) on the energy surface defined by the energy-based model.
The most difficult challenge to using the energy value directly for plausibility evaluation is that the energy-based model is a combination of uncalibrated individual experts.
To search for the local minimum, close to an initially proposed hypothesis, optimization can be used.
The goal of the optimization method is to find an optimal hypothesis $\pmb{y}^*$ and a shape $\pmb{z}^*$ that minimizes the energy-surface defined through the cost function:
\begin{equation}
    \argmin_{\pmb{y}, \pmb{z}} (\alpha_0 E_{Sil} + \alpha_1 E_{CD} + \alpha_2 E_{HoG} + \alpha_3 E_{Rot})
    \label{eqn_11}
\end{equation}
where $\alpha_n$ is a scalar value which denotes the importance factor. For our evaluation we choose two different sets of configurations for $\alpha_n$, where $C_{1}$ = [0.5,10,5,5] and $C_{2}$ = [10,0.1,1,50,$10^4$]. This configuration hyperparameter controls the impact of individual energy functions on the overall optimization time it takes to find the local minima.
An advantage of the chosen energy functions and formulations is their disposition for optimization as the individual energy functions are made differentiable.
Inspired by the requirement to find the optimal value of the hypothesis parameterized by the pose and a latent variable parameterizing the shape, we propose a novel two-step optimization to find the minima which reflects true compatibility with the observed priors.

\begin{algorithm}
\caption{Pseudocode for obtaining energy values}

\begin{algorithmic}[1]
\State \textbf{input:} hypothesis' pose params from the prediction of a base NN
\State \textbf{output:} energy value

\For{ samples in hypothesisList:}
\State Apply Threshold filtering \Comment{filter objects based on max distance from the ego\_vehicle}
\If{ checkMinMaxPoseValid(samples):}
\State fetch validBB; \Comment{Validate whether the BB is plausible or not derived from $\phi_{mean}$}
\For{validBB:}
\State check removeRadiusOutlier $\leftarrow minNbrPoints$ \Comment{For a hypothesis space there should exist a min number of points to satisfy further optimization criterias} 
\State Identify and match box from MaskRCNN scenes \Comment{False Positive when no segmentation mask is found}

\If{$(E_{HoG}$, $E_{Rot}) \leq$ minEnergyThreshold:}
\State init $\phi_{mean}$ $\leftarrow (PoseStateVector)$ \Comment{From NN}
\State costFunction$(E_{Sil},E_{CD},E_{HoG},E_{Rot})$ \Comment{apply config $C_{1}$}
\State run secondOrderOptim(costFunction, LBFGSB);
\State collect jointEnergyValue, ($y^{*}$, $z^{*}$);
\State costFunction$(E_{Sil},E_{CD},E_{HoG},E_{Rot})$ \Comment{apply config $C_{2}$}
\State run secondOrderOptim(costFunction, BFGS);
\State return  energy value w.r.t $E_{CD}$;

\Else 
\State push up the energy value quadratically; \Comment{False Positives}
\EndIf
\EndFor
\Else
\State return sqrt(calculateDeltaBetweenBB) \Comment{push up the energy value of implausible bounding box}
\EndIf
\EndFor
\end{algorithmic}
\label{Algorithm1}
\end{algorithm}


The \textbf{first optimization step,} as can be seen from line 13 of Algorithm [\ref{Algorithm1}] consists of solving a 3D rigid body pose combined with a search for the optimal shape problem.
The 3D rigid body pose of the object hypothesis can be described through a translation and rotation of the object w.r.t. the ego vehicle coordinate system.
A minimal representation of such a problem is given through 6 parameters for the 6 possible degrees-of-freedom.
However, a non-minimal representation consisting of the 3 translational components and 4 quaternion parameters is chosen for the flat euclidean spaces following \cite{Blanco2010}.
The pose state vector is therefore given as:
\begin{equation}
    \pmb{\tau} = \pmb{\tau}_O^{EV} = [\pmb{t}_O^{EV}, \pmb{q}_O^{EV}] =
    \begin{bmatrix}t_x & t_y & t_z & q_w & q_x & q_y & q_z \end{bmatrix} \in \mathbb{R}^7
\end{equation}
where, $\pmb{t}_O^{EV} = \begin{pmatrix} t_x & t_y & t_z \end{pmatrix}^\top$ are the components of the translation vector describing the shift of points from the object-centered coordinate system (O) to the ego-vehicle-centered coordinate system (EV) and $\pmb{q}_O^{EV} = q_w + q_x \mathcal{i} + q_y \mathcal{j} + q_z \mathcal{k}$ describes the quaternion parameters of the transformation.
For the shape parameters $z_n$, the 5 shape weights corresponding to the 5 first primary components of the shape manifold are chosen.
The state vector is therefore a concatenation of the pose and shape parameters:
\begin{equation}
    \pmb{\xi} = [\pmb{\tau}_O^{EV}; \pmb{z}] =
    \bigl[ t_x \quad t_y \quad t_z \quad q_w \quad q_x
    \quad q_y \quad q_z \quad z_0 \quad z_1 \quad z_2 \quad z_3 \quad z_4 \bigr]
    \in \mathbb{R}^{12}
\end{equation}
Each shape weight is bounded to an interval of $[-1, 1]$. The bounded version of the $2^{nd}$ order quasi-newton optimization method, L-BFGSB \cite{Nocedal2006}, is chosen for the current optimization step. A lightweight optimization method is chosen, as the objective of this step is to find the optimal shape parameter which fits the point cloud observation. The optimization step uses $C_{1}$ configuration as a design choice. The weight vector of the configuration expresses relatively a higher weight towards $E_{CD}$, as this pushes the energy function to find a better shape to fit to the observed point cloud, which otherwise suffers from translation and rotation errors.
The $E_{Sil}$ helps to attain a faster optimization by guiding the energy function towards minima. The optimized ($\pmb{y}^*$,$\pmb{z}^*$) is the best fit to the proposal from the NN.  
The resulting energy from this step is in itself sufficient to argue against the plausibility of the object. But there exist cases similar to Fig. \ref{fig:misclassification_truck} where a misclassification with objects from other classes could provide a false fit, contributing to False Negatives situations. To mitigate such cases, a second optimization is needed.  

The \textbf{second optimization step}, as can be seen from line 16 of Algorithm [\ref{Algorithm1}] uses $C_{2}$ configuration parameters with BFGS optimization method. The goal of this optimization step is to collect the energy values given the strong requirements for the object to be on the ground while the wheels touching the road surface capture through $E_{HoG}$ and $E_{Rot}$. This design choice yields us an additional check to monitor whether there exists a significant gradient change in $E_{CD}$ while the overall function pushes the rest of the priors towards minima. For TP detections, this optimization step would result in a non-significant gradient change to the energy value, as the function has already attained local minima on the energy surface. In case of an FP, arising out of misclassification, e.g., as shown in Fig. \ref{fig:misclassification_truck} where a truck is classified with high probability of being a car, the gradient change w.r.t to $E_{CD}$ should be positive indicating an incompatibility between the priors and the observed hypothesis. 
In both optimization steps, the Jacobian, and the Hessian, are numerically approximated to obtain better results.  

\subsection{Metric}
At the end of the optimization process, each proposal in the queue to be verified receives an associated uncalibrated energy value. Due to the uncalibrated nature of the energy values, a decision rule based on a threshold over the energy value needs to be chosen. To search for the cutoff threshold, we created a synthetic dataset from NuScenes Validation dataset~\cite{nuscenes}. The synthetic dataset contains a balanced mix of TP and FP created by random perturbation to the bounding boxes, including some boxes being pushed higher from the ground, simulating an adversarial attack. The baseline has a probability value of 0.5 as our perturbed dataset is balanced.
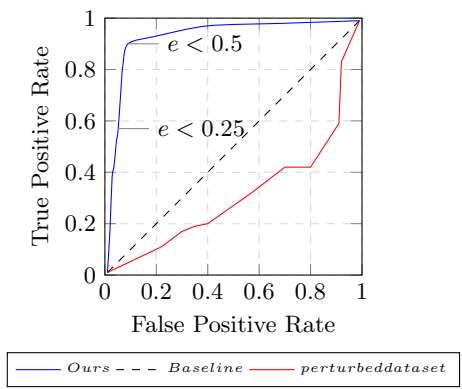
\begin{figure}[h!]
\centering
\begin{tikzpicture}
  \begin{axis}[
      width=5cm,height=5cm,
      grid=major, 
      grid style={dashed,gray!30},
      xmin=0,ymin=0,
      xmax=1,ymax=1,
      xlabel=False Positive Rate, 
      ylabel=True Positive Rate,
      xtick={0,.2,.4,.6,.8,1},
      ytick={0,.2,.4,.6,.8,1},
      legend  style={font=\tiny,at={(0.5 ,-0.30)}, anchor=north,legend  columns =-1},
    ]
    \addplot+
        [smooth] [mark=none]
        table[x=FPR,y=TPR,col sep=comma] {graphics/data/roc.csv}; 
        \node[coordinate ,pin=0:{$e<0.5$}] at (axis cs:0.09,0.90) {};
        \node[coordinate ,pin=0:{$e<0.25$}] at (axis cs:0.052,0.57) {};
        \addlegendentry{$Ours$}
    \addplot[mark=none,dashed,color=black] coordinates {
	    (0.01,0.01)
	    (0.25,0.25)
	    (0.5,0.5)
	    (0.75,0.75)
	    (0.99,0.99)
    };
        \addlegendentry{$Baseline$}
    \addplot+ [mark=none]
        [color=red] coordinates {
        (0.01,0.01)
        (0.22,0.11)
        (0.3,0.17)
        (0.35,0.19)
        (0.4,0.2)
        (0.57,0.32)
        (0.7,0.42)
        (0.8,0.42)
        (0.91,0.59)
        (0.92,0.83)
        (0.99,0.99)
    };  
        \addlegendentry{$perturbed dataset$}
  \end{axis}
\end{tikzpicture}
\caption{Quantitative evaluation of the synthetic dataset to find an optimal energy threshold to act as a decision rule-based filter.}
\label{plot:roc_graph}
\end{figure}
\begin{figure}[h]
\centering
    \begin{tikzpicture}[baseline]
    \begin{axis}[
          width=5cm, height=5cm,
          grid=major, 
          grid style={dashed,gray!30},
          xmin=0,ymin=0,
          xmax=1,ymax=1,
          xlabel=Recall, 
          ylabel=Precision,
          xtick={0,.2,.4,.6,.8,1},
          ytick={0,.2,.4,.6,.8,1},
          legend  style={font=\tiny, at={(0.5 ,-0.30)}, anchor=north,legend  columns =-1},
    ]

    \addplot [mark=none,color=black] coordinates{(0,1) (0.05,0.64) (0.07,0.62)
                        (0.18,0.43) (0.3,0.38) (0.4,0)
                         (1.0,0.0)};
        \addlegendentry{\emph{MonoRUn}}
    \addplot [mark=none,color=blue] coordinates{(0,0.93) (0.17,0.92) (0.4,0.87) 
                        (0.5,0.90) (0.7,0.88) (0.82,0.79) 
                        (0.83,0.60) (0.86,0.0)};
        \addlegendentry{\emph{MonoRUn+Ours}}
    \end{axis}
    \end{tikzpicture}
    ~
    \begin{tikzpicture}[baseline]
    \begin{axis}[
          width=5cm, height=5cm,
          grid=major, 
          grid style={dashed,gray!30},
          xmin=0,ymin=0,
          xmax=1,ymax=1,
          xlabel=Recall, 
          ylabel=Precision,
          xtick={0,.2,.4,.6,.8,1},
          ytick={0,.2,.4,.6,.8,1},
          legend  style={font=\tiny, at={(0.5 ,-0.30)}, anchor=north,legend  columns =-1},
    ]
    \addplot [mark=none,color=red] coordinates{(0,0.78) (0.09,0.7) (0.14,0.63)
                        (0.19,0.58) (0.385,0.39) (0.42,0.2)
                        (0.43,0.0) (1.0,0.0)};
        \addlegendentry{\emph{MonoRUn+LiDAR}}
    \addplot [mark=none,color=blue] coordinates{(0,0.90) (0.17,0.93) (0.4,0.91) 
                        (0.5,0.90) (0.7,0.88) (0.82,0.82) 
                        (0.83,0.58) (0.9,0.0)};
        \addlegendentry{\emph{MonoRUn+Ours}}
    \end{axis}
    \end{tikzpicture}
    \caption{Quantitative evaluation of MonoRUn network and ours on KITTI easy evaluation. The left chart shows the baseline for the network which uses no LiDAR supervision when compared with the right}
    \label{plot:PR_chart}

\end{figure}
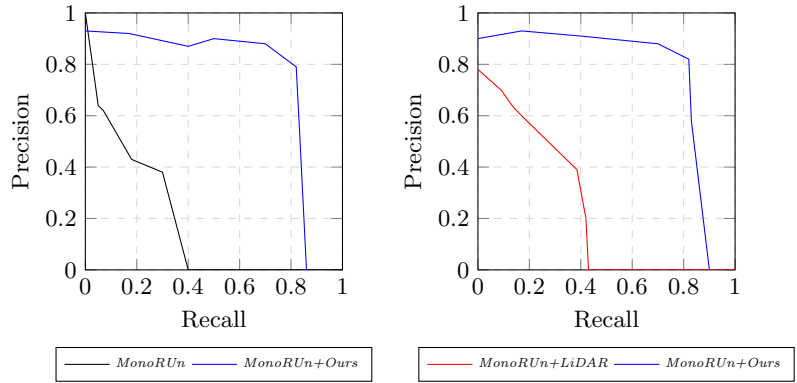
The red line on the ROC chart Fig. \ref{plot:roc_graph} represents the amount of FP for different IOU thresholds. Each energy values contribute to the plausibility $pl(\pmb{y}) \in \{1, 0\}$ and is evaluated to be either plausible or implausible based on an empirical threshold $\kappa$ as defined by the following equation. 
\begin{equation}
    pl(\pmb{y}) = \begin{cases}1, & E(\pmb{y}) \leq \kappa \\ 0, & E(\pmb{y}) > \kappa \end{cases}
\end{equation} 
Our plausibility verification was then done for each of the samples on this perturbed dataset, and the results can be seen from the graph.
The blue line from the graph is obtained for different thresholds' of energy values. Do note that the energies defined are uncalibrated. TPR has the interesting property of capturing the impact of TNs and FPs. When FPs are detected, this classification gets converted to TNs leading to a shift towards the left side of the graph. An energy threshold value of 0.5 is chosen, since this threshold converts most of the FPs present within the synthetic dataset to TN.

\begin{figure}[h!]
\centering
  \begin{subfigure}[b]{0.30\textwidth}
  \centering
    \includegraphics[width=0.95\linewidth]{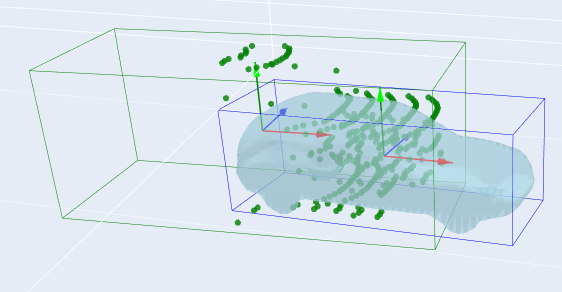}
    \caption{Back View }
  \end{subfigure}
  \begin{subfigure}[b]{0.30\textwidth}
  \centering
    \includegraphics[width=0.95\linewidth]{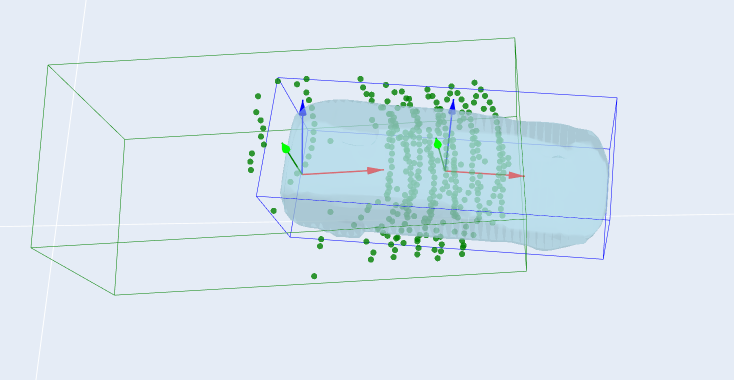}
    \caption{Top View}
  \end{subfigure}
  \begin{subfigure}[b]{0.30\textwidth}
  \centering
    \includegraphics[width=0.95\linewidth]{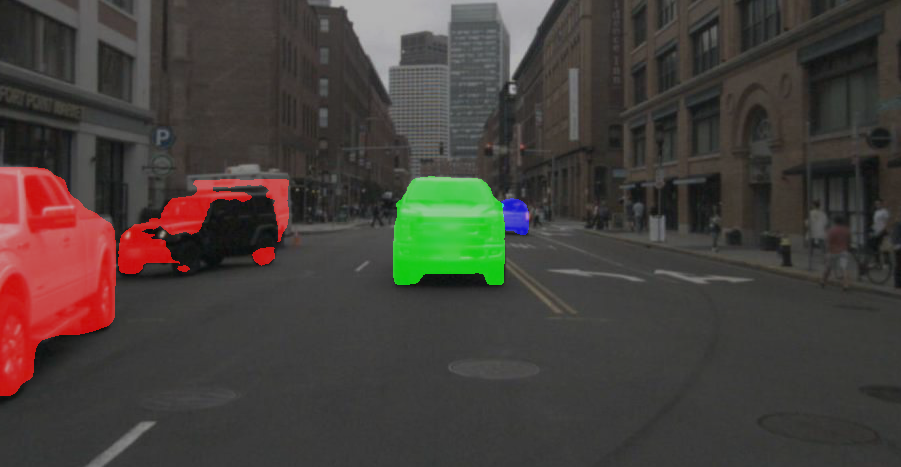}
    \caption{Segmentation mask}
  \end{subfigure}
  \caption{Quantitative evaluation of misclassification. In this case, the truck was misclassified as a car and point cloud observations doesn't fit the optimized shape leading to a high-energy value. (a) and (b) shows the back view and top view of the sample. Green BB, ground truth proposal. Blue BB = Optimized pose and shape vector (c) shows the segmentation mask from MaskRCNN which was trained on cityscapes to exhibit worst-case performance}
  \label{fig:misclassification_truck}
\end{figure}

\begin{figure}[ht] 
  \begin{subfigure}[b]{0.25\textwidth}
    \centering
    \includegraphics[width=0.90\textwidth]{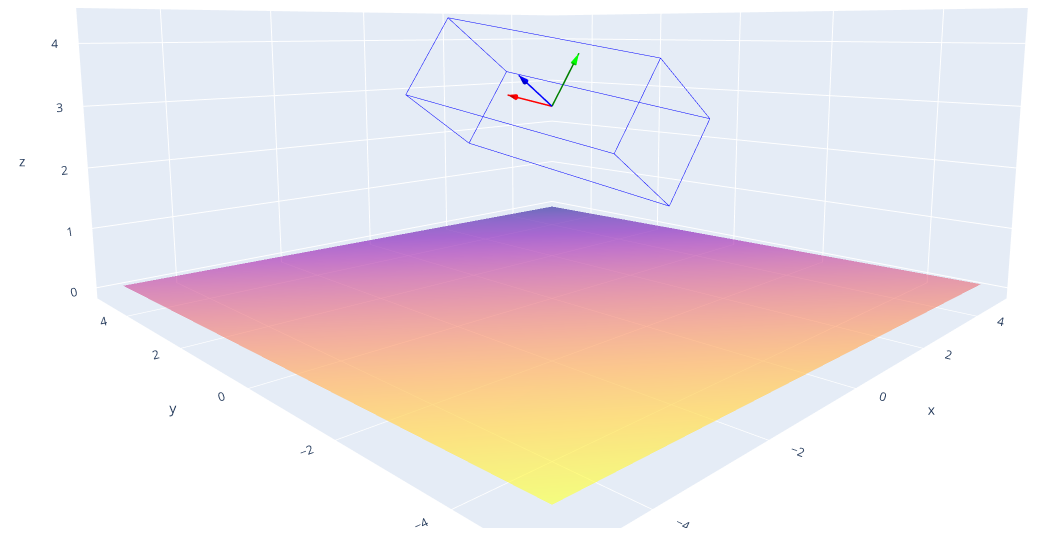}
    \caption{$E=E_{HoG}+E_{Rot}\\=4.81+3.4=8.21$}
    \label{fig:gp_optim_1}
  \end{subfigure}
  \begin{subfigure}[b]{0.25\textwidth}
    \centering
    \includegraphics[width=0.90\textwidth]{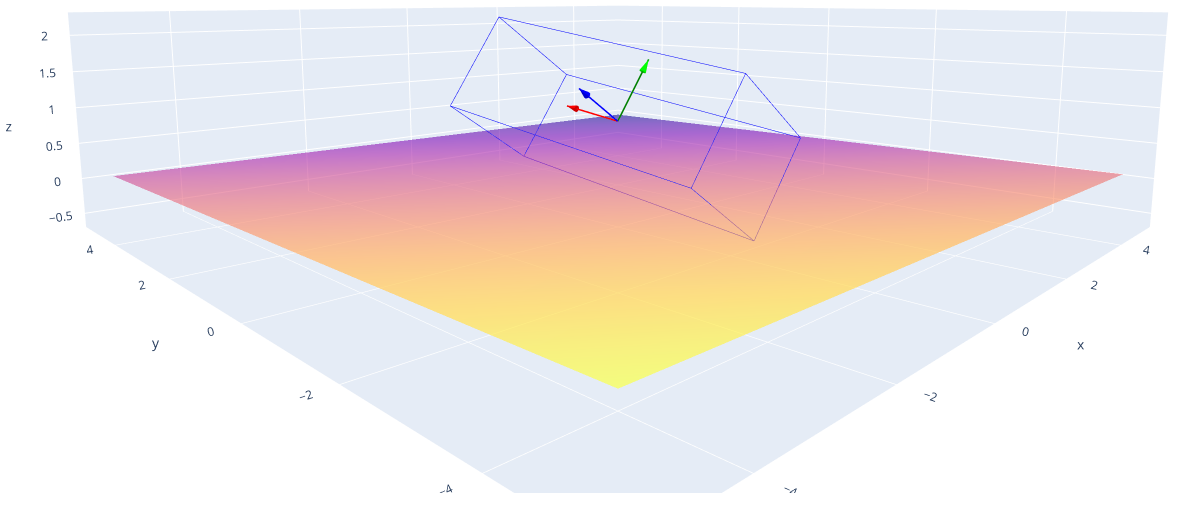} 
    \caption{$E=E_{HoG}^*+E_{Rot}\\=0+3.4=3.4$} 
    \label{fig:gp_optim_2}
  \end{subfigure} 
  \begin{subfigure}[b]{0.25\textwidth}
    \centering
    \includegraphics[width=0.90\textwidth]{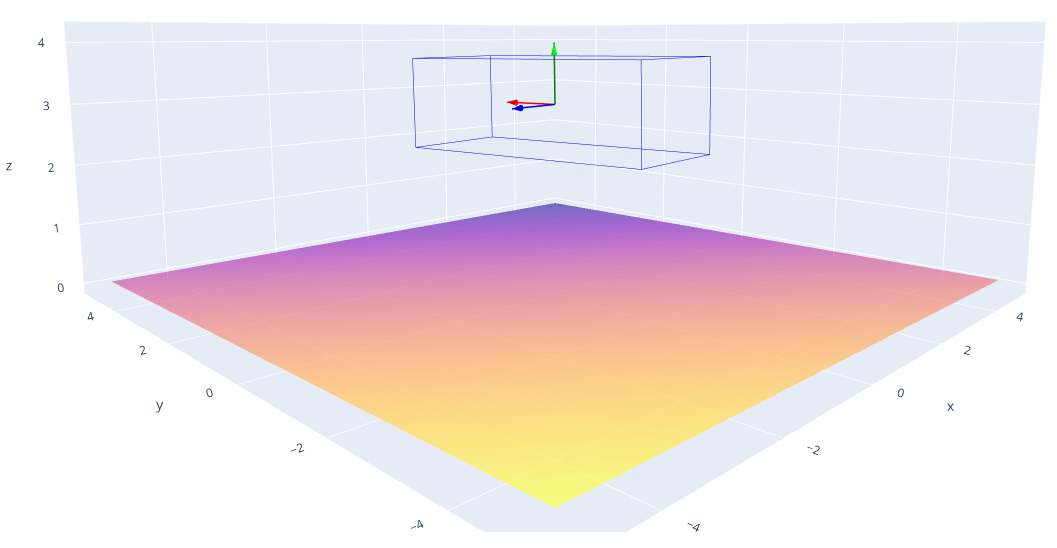} 
    \caption{$E=E_{HoG}+E_{Rot}^*\\=4.81+0=4.81$} 
    \label{fig:gp_optim_3}
  \end{subfigure}
  \begin{subfigure}[b]{0.25\textwidth}
    \centering
    \includegraphics[width=0.90\textwidth]{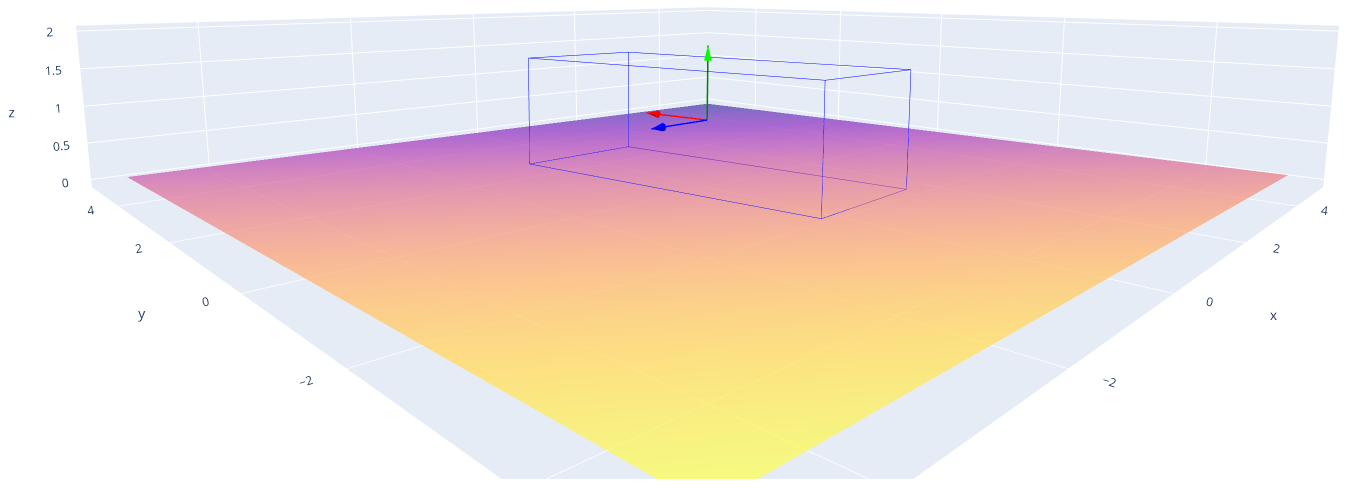} 
    \caption{$E^*=E_{HoG}^*+E_{Rot}^*\\=0$} 
    \label{fig:gp_optim_4}
  \end{subfigure} 
  \caption{Qualitative example of Height-over-Ground and Rotation energy prior-based optimization for a given ground plane and a hypothesis (blue bounding box) with random position and orientation, as shown in (a). The result of optimizing the priors individually and jointly is shown in (b), (c) and (d).}
  \label{fig:gp_optim} 
\end{figure}
\begin{figure}[ht] 
  \begin{subfigure}[b]{0.25\textwidth}
    \includegraphics[width=0.95\textwidth]{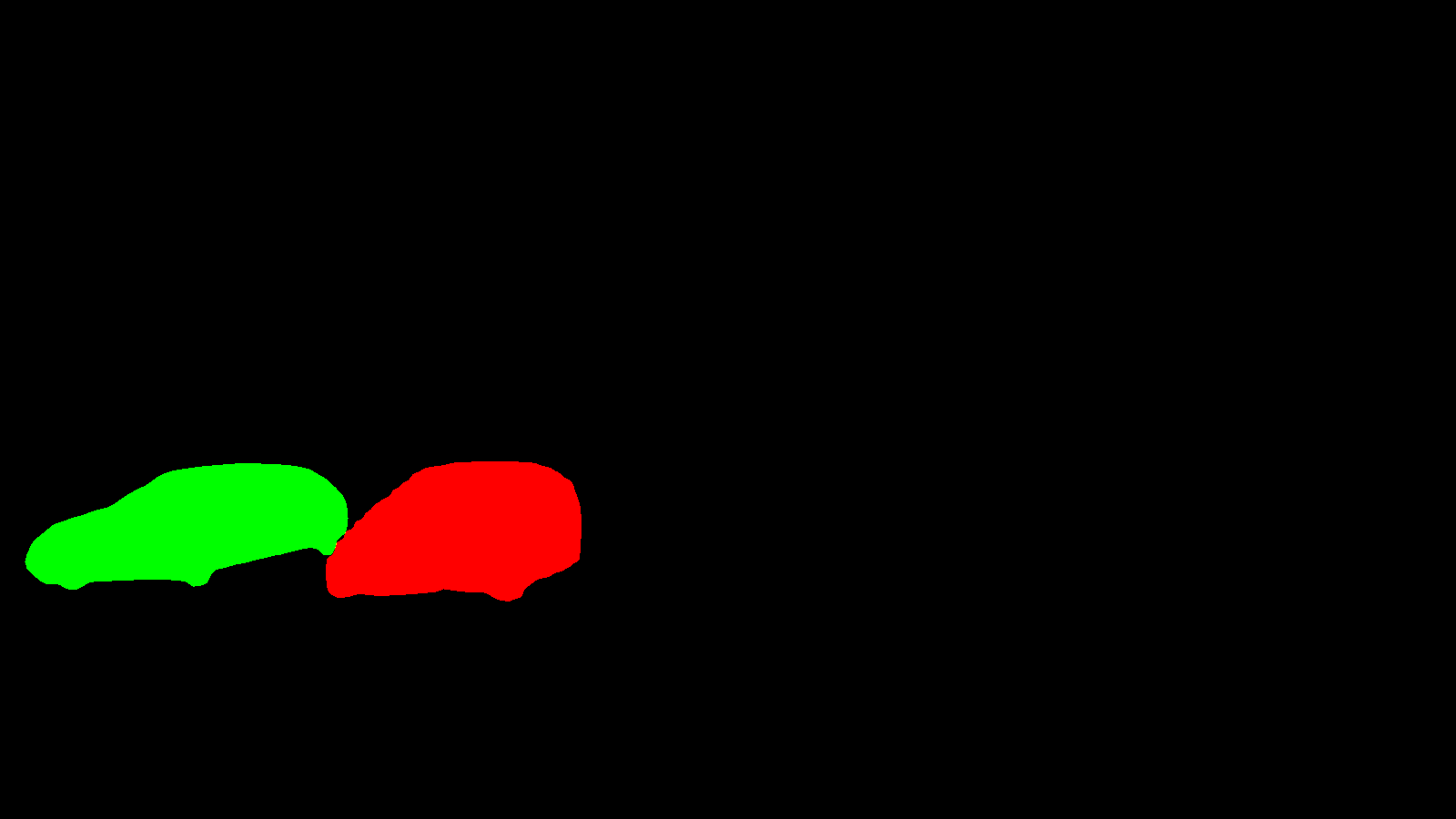} 
    \caption{$E_{Sil}(\pmb{y}_0) = 0.6$} 
    \label{a}
  \end{subfigure}
  \begin{subfigure}[b]{0.25\textwidth}
    \includegraphics[width=0.95\textwidth]{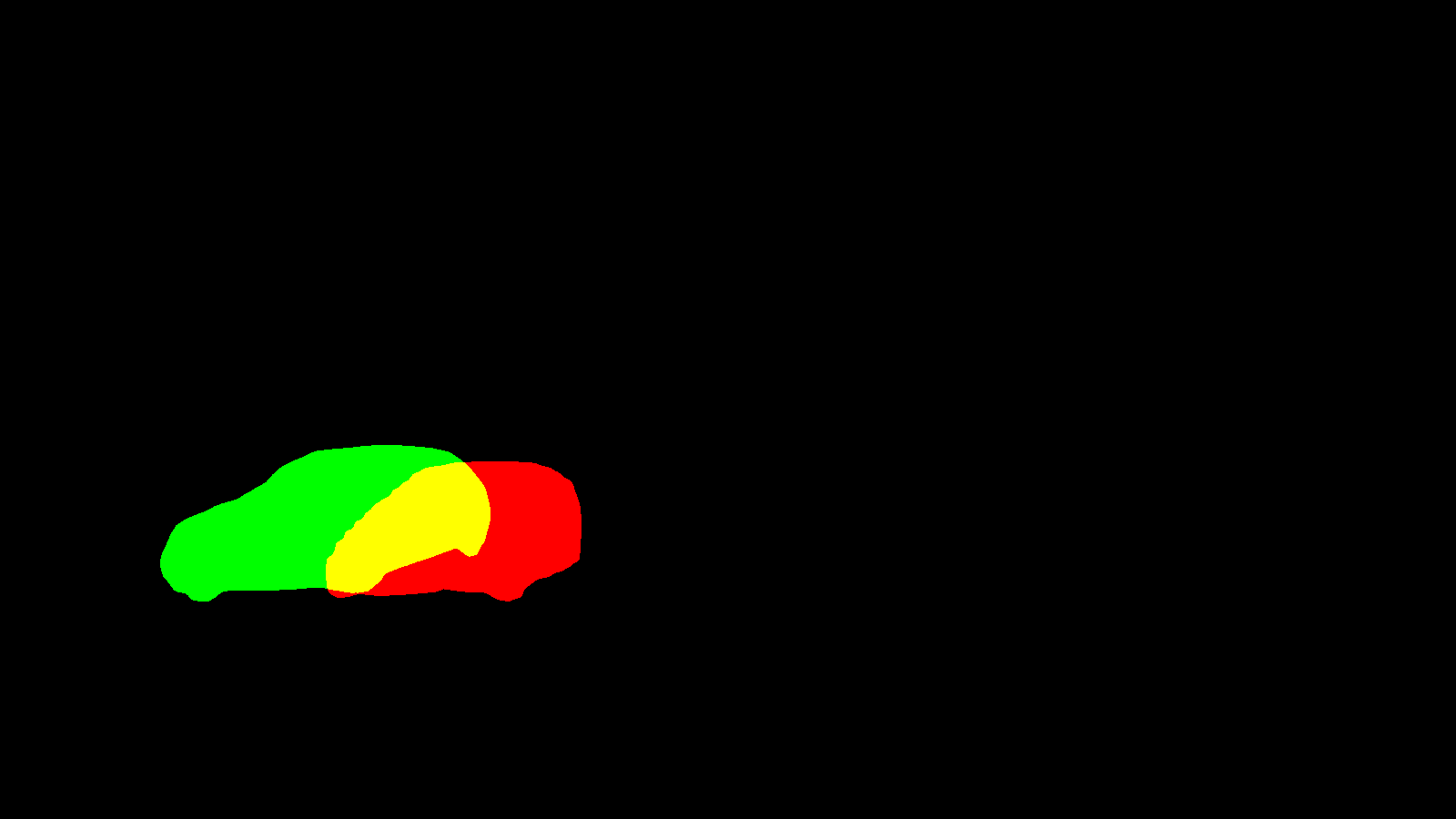} 
    \caption{$E_{Sil}(\pmb{y}_1) = 0.4$} 
    \label{b}
  \end{subfigure} 
  \begin{subfigure}[b]{0.25\textwidth}
    \includegraphics[width=0.95\textwidth]{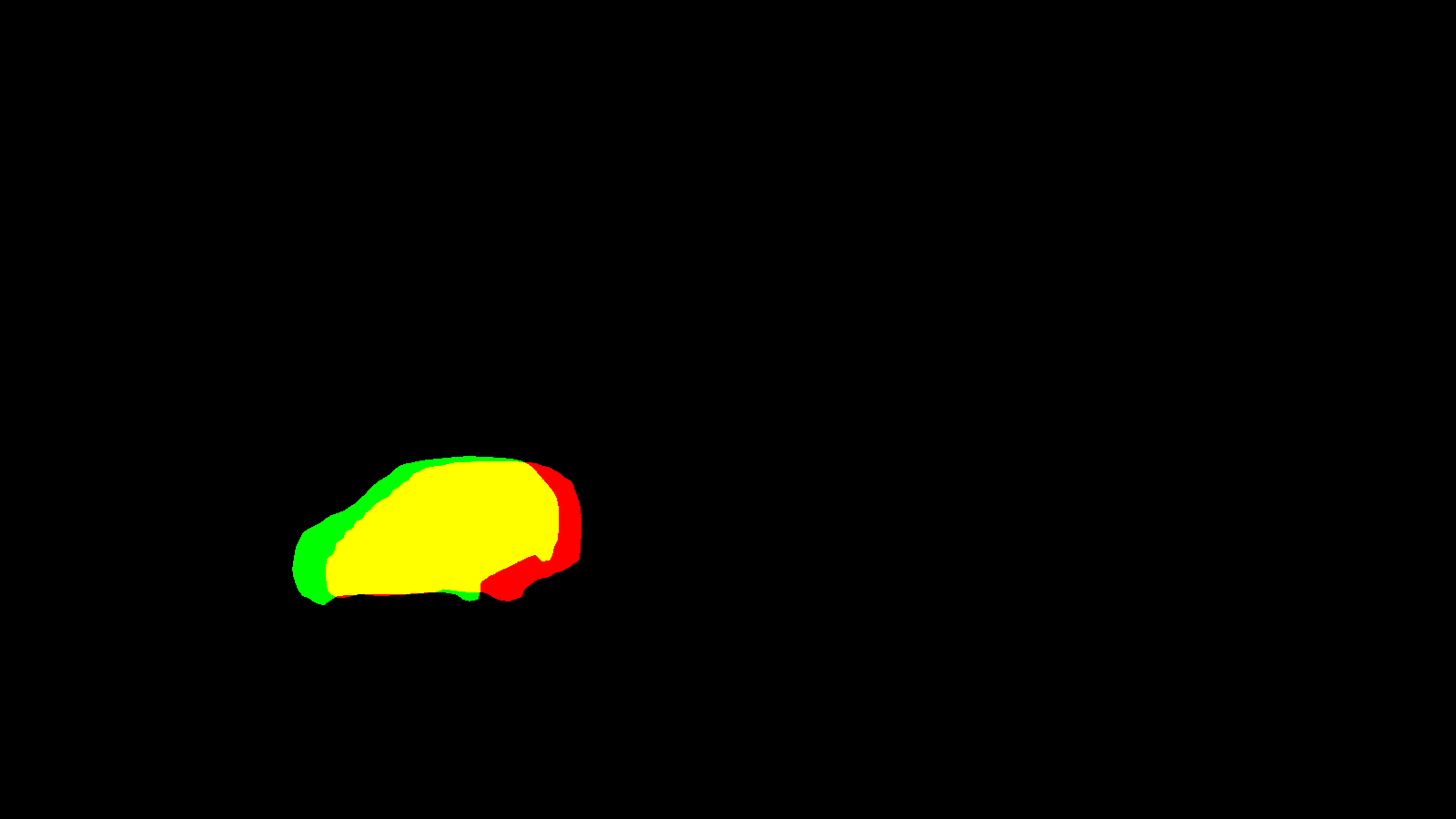} 
    \caption{$E_{Sil}(\pmb{y}_2) = 0.2$} 
    \label{c}
  \end{subfigure}
  \begin{subfigure}[b]{0.25\textwidth}
    \includegraphics[width=0.95\textwidth]{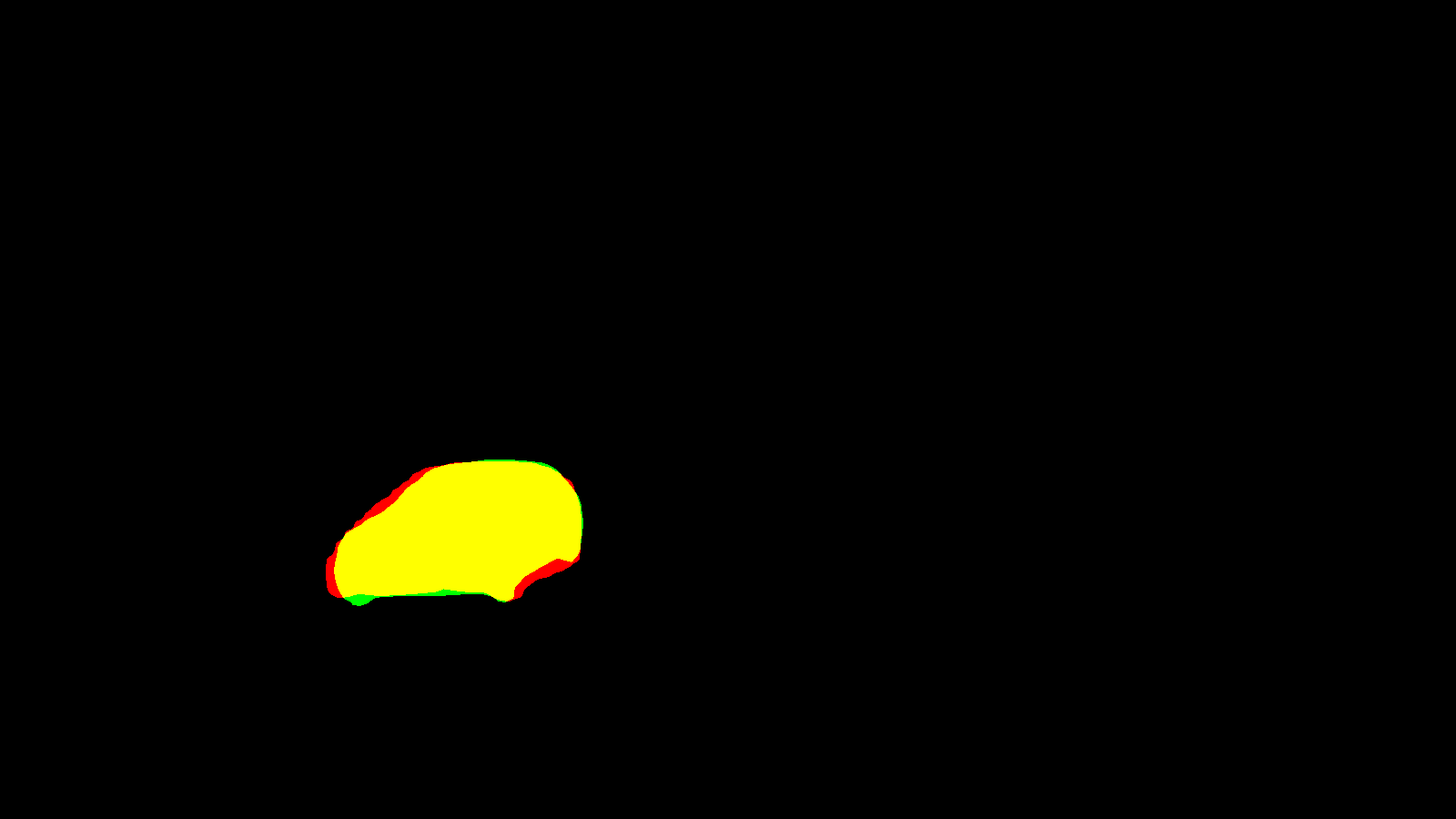} 
    \caption{$E_{Sil}(\pmb{y}_3) = 0.01$} 
    \label{d}
  \end{subfigure} 
  \caption{ Depicted are the overlap (yellow) of a segmentation mask (red) obtained from a NuScenes sample and a silhouette rendering (green) of the mean shape prior of an initial hypothesis, the optimized hypothesis and intermediate steps of the optimization process.}
  \label{fig:Figure7} 
\end{figure}

\begin{figure}[ht]
\centering
    \begin{subfigure}[b]{0.40\textwidth}
	    \centering
        \includegraphics[width=0.95\columnwidth]{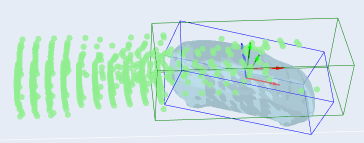}
        \caption{Motion artifacts}
        \label{fig:limit-motion}
    \end{subfigure}
    \begin{subfigure}[b]{0.40\textwidth}
	    \centering
        \includegraphics[width=0.95\columnwidth]{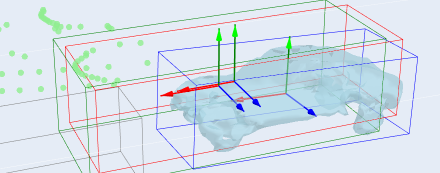}
        \caption{Breaking Shape}
        \label{fig:limit-shape}
    \end{subfigure}
    \caption{Failure cases: True positives are being misclassified as False positives due to sensor noise. Green box represents the ground truth, blue box represents the optimized shape while red represents the initial hypothesis}
    \label{fig:limit-more}
\end{figure}
\section{Experiments}
As our work focuses towards reducing false positives and acts as a parallel module which verifies proposals from the NN we need to choose a model which suits the evaluation criteria. Our module requires no training components besides the MaskRCNN module, which needs to be retrained for the ODD. To justify the capabilities of our parallel module, we choose MonoRUn as our base network and measure its performance with the KITTI \cite{Geiger2012CVPR} test dataset.  
For each of the hypothesis/proposals in the scene, we have an initial plausibility check against Height-over-Ground and Rotation consistent priors. If the 3DOP produces a high-energy value (E$>$0.5) we attribute the hypothesis as a \textit{false positive}. In addition to the height and rotation checks, we also filter the proposals based on a distance threshold to limit our verification space and this is chosen as 30 m in front and back, 15 m to the left and right of the ego vehicle's camera position. 
We also apply checks based on the number of points in the LiDAR space and remove proposals which have implausible Bounding box shapes. These simple checks negate the need to validate the hypothesis against further energy values, saving valuable computation time. The implication stays consistent with our world view (object belonging to class car can't float in free space) about an object's position, w.r.t our prior knowledge. 

In the Fig. \ref{plot:PR_chart} we qualitatively evaluated our energy-based threshold filter against MonoRUn network with and without the LiDAR supervision. The AP of our filter exceeds the base NN by a wide margin. From the plots, the precision, which measure the total positivity of the observed samples, seems to achieve moderately high value which showcases that our method is effective in reducing the false positives while maintaining a significantly reduced False Negatives relative to the base NN. 

In Fig. \ref{fig:gp_optim} for a given ground estimate, the height-over-ground and rotation energy prior can be calculated. Fig. \ref{fig:gp_optim} shows one of the samples which were randomly shifted and rotated. Fig. \ref{fig:gp_optim_1} shows the initial hypothesis, floating above the ground plane, with the orientation being not aligned with the ground plane.This deviation from the requirements are encoded into the energy priors (e.g., a car should touch the ground with its four wheels) and is reflected in the energy values for the particular sample.
Given a segmentation mask and a rendered silhouette mask of the optimal hypothesis, the $E_{Sil}$ evaluates the compatibility of the two. For this experiment, the differentiable renderer was set to a down sample factor of 8, meaning that only every 64th (82) pixel the projection function was evaluated. This downsampling helps to reducing the computational intensity whilst preserving the silhouette details. Along each ray, points were sampled every 0.3 m to a maximum distance of 30 m. The parameter $\xi$, controlling the steepness of the sigmoid function in the projection function, was chosen to be \textminus25.
Fig. \ref{fig:Figure7} section (a) shows how the initial hypothesis (green mask) barely coincides with the segmentation mask (obtained from MaskRCNN) leading to an $E_{Sil}(y_0)=0.6$. As the energy is the mean over all the residual values of each pixel, describing the agreement with values in the interval $[0,1]$, the energy function is bounded $E_{Sil} \in [0,1]$.
During optimization, as shown in Fig. \ref{b} and \ref{c}, one can see that the energy decreases, leading to an optimal energy value in Fig. \ref{d}.
During our experiments we found some failure cases as can be seen from Fig. \ref{fig:limit-more} being cause due to motion artifacts and sensor noise from the LiDAR.

Computation time required for the optimization schema as measured on Intel-Core-i7 comes around at an average of 680ms per sample without any parallelization. $70\%$ of the compute time is spent for the optimization of the shape weights in $E_{CD}$. Since CD is a pointwise comparison between 2 different point clouds, this computation time could significantly be reduced by pushing the distance estimation to a GPU. 

\section{Conclusion}
We propose a new schema which functions as a parallel checker module to verify the predictions from a 3D Object Detector. The different energy function we propose utilizes cross sensor data flows with simple priors to validate an object. We also demonstrated the viability of a decision rule base threshold filter through the synthetic dataset. From our experiments, we extensively showcase the viability of the two-step optimization schema and modified renderer towards effectively utilizing the priors in reducing the amount of \textit{false positives} relative to the proposals. In our future work, we are planning to use motion priors along with map information to further argue about the validity of an object in space.\

\subsection*{Acknowledgements} The research leading to these results is funded by the German Federal Ministry for Economic Affairs and Climate Action within the project “KI Wissen". The authors would like to thank the consortium for the successful cooperation.

\bibliographystyle{splncs04}
\bibliography{references}
\end{document}